\documentclass{article}

\usepackage{PRIMEarxiv}

\usepackage[utf8]{inputenc} 
\usepackage[T1]{fontenc}    
\usepackage{hyperref}       
\usepackage{url}            
\usepackage{booktabs}       
\usepackage{amsfonts}       
\usepackage{nicefrac}       
\usepackage{microtype}      
\usepackage{lipsum}
\usepackage{fancyhdr}       
\usepackage{graphicx}       
\graphicspath{{media/}}     
\usepackage{subcaption}
\usepackage{amsmath}

\title{Multi-Label Continual Learning for the Medical Domain: A Novel Benchmark}

\author{
  Marina Ceccon \\
  University of Padova \\
  Padova\\
  \texttt{marina.ceccon@phd.unipd.it} \\
   \And
  Davide Dalle Pezze \\
  University of Padova \\
  Padova\\
  \texttt{davide.dallepezze@unipd.it} \\
  \And
  Alessandro Fabris \\
  Max Planck Institute for Security and Privacy \\
  Bochum\\
  \texttt{alessandro.fabris@mpi-sp.org} \\
  \And
  Gian Antonio Susto \\
  University of Padova \\
  Padova\\
  \texttt{gianantonio.susto@unipd.it} \\
}

\begin{document}
\maketitle

\begin{abstract}
Despite the critical importance of the medical domain in Deep Learning, most of the research in this area solely focuses on training models in static environments. It is only in recent years that research has begun to address dynamic environments and tackle the Catastrophic Forgetting problem through Continual Learning (CL) techniques.
Previous studies have primarily focused on scenarios such as Domain Incremental Learning and Class Incremental Learning, which do not fully capture the complexity of real-world applications. 
Therefore, in this work, we propose a novel benchmark combining the challenges of new class arrivals and domain shifts in a single framework, by considering the New Instances and New Classes (NIC) scenario. 
This benchmark aims to model a realistic CL setting for the multi-label classification problem in medical imaging. Additionally, it encompasses a greater number of tasks compared to previously tested scenarios.
Specifically, our benchmark consists of two datasets (NIH and CXP), nineteen classes, and seven tasks, a stream longer than the previously tested ones.
To solve common challenges (e.g., the task inference problem) found in the CIL and NIC scenarios, we propose a novel approach called Replay Consolidation with Label Propagation (RCLP).
Our method surpasses existing approaches, exhibiting superior performance with minimal forgetting.
\end{abstract}


\begin{figure*}[!t]
   \centering
   \includegraphics[width=\linewidth, trim = 0 0 0 0]{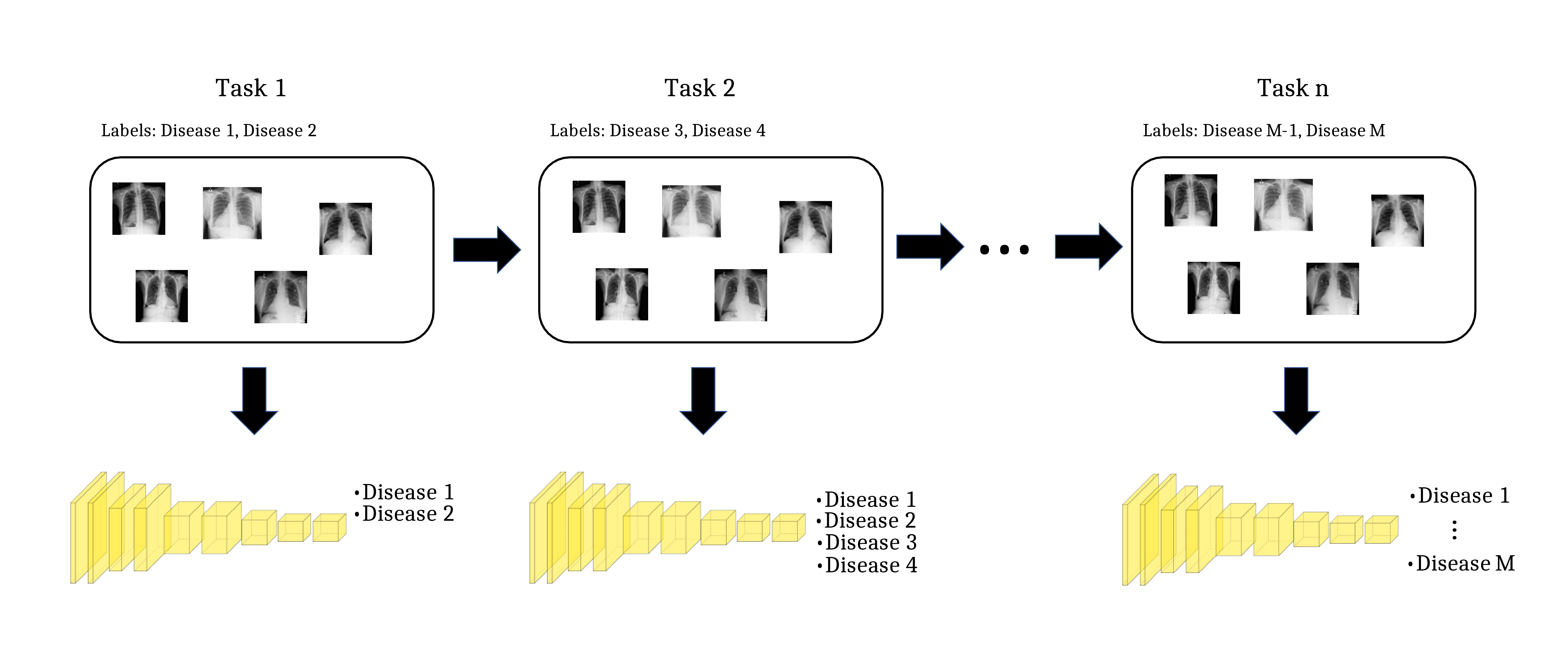}
   \caption{Scheme of the multi-label CL setting in the context of classification of chest X-rays. Diagnostic capabilities are expanded over time with new tasks.}
   \label{fig:multiLabelScheme}
\end{figure*}

\section{Introduction}
\label{sec:intro}

In recent years, several studies have proven the efficacy of using Deep Learning models to detect diseases from chest X-ray images \cite{lenga2020continual, zhang2022improving}.

While these models allowed for improved state-of-the-art in the medical field, several significant challenges must be addressed before considering their use to support decision-making in realistic scenarios.
One such challenge is ensuring the model's adaptability in dynamic environments, where shifts in input data distribution may occur over time \cite{kumari2023continual, srivastava2021continual}.

For instance, each hospital may employ machines with different image acquisition techniques; this can perturb the performance of a previously trained model, requiring ongoing recalibration and refinement \cite{lenga2020continual}.
Moreover, the need to detect diseases not initially incorporated into the trained model may arise. 
For instance, the labeling of additional classes may have been provided at a later point in time, reflecting evolving medical understanding and diagnostic criteria \cite{singh2023class}. 

Failure to address these challenges can severely limit the model's utility and constrain its ability to effectively serve patient needs.

A paradigm called Continual Learning (CL) has emerged in the literature to solve this problem. CL enables the model to adapt to new data while retaining knowledge from the old data.

Previous studies in this domain focused on solving the problem in two different settings: Domain Incremental Learning (DIL) and Class Incremental Learning (CIL). 

However, these scenarios fail to accurately represent real-world challenges where both new classes and domain shifts can happen. 
Therefore, we propose a novel benchmark for the medical imaging field based on the New Instances \& New Classes (NIC) scenario \cite{lomonaco2017core50}. 
Specifically, we evaluate this scenario in the context of pathology classification of chest X-ray images, considering nineteen classes, seven tasks, and two domains.
This setting combines the challenges of both new class arrivals and domain shifts within a single framework, mirroring the complexities often encountered in realistic applications like medical imaging \cite{lomonaco2017core50}. 

On the proposed benchmark, the most well-known techniques in the CL field are tested.
In particular, replay-based and distillation-based approaches are tested since they are often used to solve the CIL scenario in image classification \cite{singh2023class,kumari2023continual}. 
However, these approaches face limitations in multi-label settings such as the one considered.
Replay-based approaches may suffer from task interference between the samples from the current task and the replayed samples \cite{ul2021incremental}. 
Instead, distillation-based methods deteriorate when old classes do not reappear in future tasks \cite{9575493}.
Therefore, we introduce a novel approach called Replay Consolidation with Label Propagation (RCLP) to solve these challenges.

Overall, we make the following contributions:
\begin{itemize}
    \item We introduce a novel benchmark for CL in medical imaging, combining the challenges of new class arrivals and domain shifts in a single framework. 
    \item We propose Replay Consolidation with Label Propagation, a novel method to address the multi-label image classification problem in the medical imaging setting.
    \item Experimental results demonstrate the effectiveness of RCLP, outperforming existing methods and achieving performance improvements. 
\end{itemize}
In addition, to promote further research in the domain and facilitate the advancement of novel methodologies along with comparisons with state-of-the-art approaches, we make the code available.\footnote{ \url{ https://anonymous.4open.science/r/RCLP-CFC7/README.md}}
\\
Our work is structured as follows. 
Sec.~\ref{sec:related_work} provides an overview of the CL literature with a focus on the medical domain.
Sec.~\ref{sec:our_scenario} presents the proposed benchmark for multi-label medical imaging, while Sec.~\ref{sec:our_approach} proposes the RCLP approach. 
In Sec.~\ref{sec:experimental_setting} describes the experimental setting, and in Sec.~\ref{sec:results} reports the results of our experiments. Lastly, Sec.~\ref{sec:conclusions_future_work} concludes this work by discussing limitations and future research directions.

\begin{figure*}[!t]
   \centering
   \includegraphics[width=\linewidth, trim = 0 0 0 0]{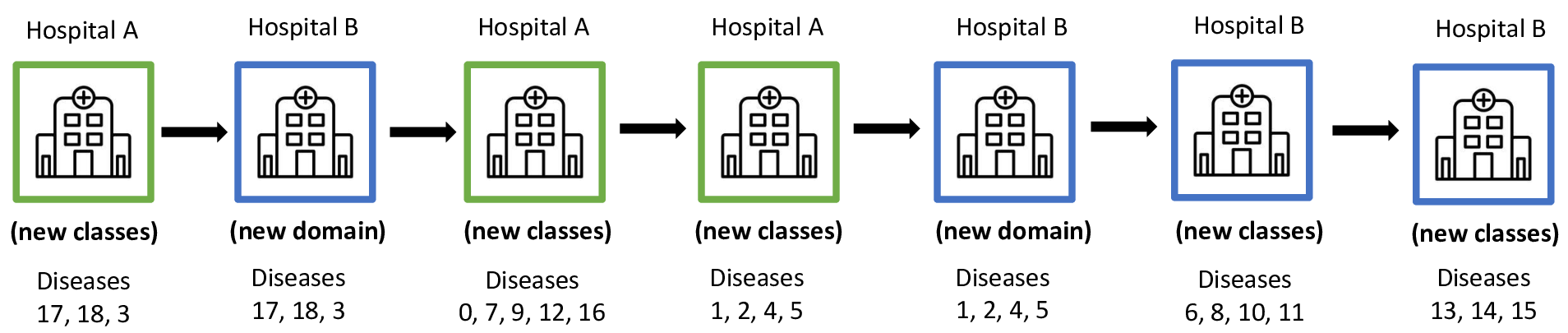}
   \caption{ Benchmark proposed for the medical imaging field. The figure presents a New Instances \& New Classes scenario, where some tasks introduce new classes while other tasks involve previously seen classes but with a shift in the input data distribution. Our proposed stream consists of a sequence of seven tasks, encompassing a total of nineteen classes across two domains. }
   \label{fig:nic_scenario}
\end{figure*}

\section{Related Work}
\label{sec:related_work}
In Sec. \ref{subsec:intro_cl}, we provide an overview of the most popular CL scenarios and methods.
Following that, we focus on the medical domain and chest X-ray image classification for the CL setting in Sec. \ref{subsec:intro_medical_cl}.

\subsection{Continual Learning}
\label{subsec:intro_cl}
CL has emerged as a framework to adapt models to new data distributions without forgetting what they have already learned \cite{lesort2019continual}. 
The main problem tackled by the CL methods is the so-called \emph{catastrophic forgetting}.
By fine-tuning the new distributions of data, the knowledge of old data gets completely corrupted \cite{kirkpatrick2017overcoming, Replaypaper}.
In the CL literature, most works refer to three CL scenarios: Domain-Incremental, Class-Incremental, and Task-Incremental Learning \cite{vandevan2019}.
In the DIL scenario, each new task presents a shift in the input data distribution but the classes are always the same \cite{lesort2019continual}.
In the CIL scenario, each task contains new classes.
The Task-Incremental scenario, contrary to the previous two, has the peculiarity of providing the task ID during the testing phase.

The methods in the CL literature can be grouped into three big families.
\textit{Rehearsal-based} techniques assume storing and reusing past data samples during training. 
Within this category, various approaches exist; a prominent method is Experience Replay \cite{rolnick2019experience}, simply referred to as \textit{Replay}.
This method stores a portion of the old data in memory. While training on a new task, the new data is combined with the data from the memory buffer to maintain knowledge of old tasks.
\\
\textit{Regularization-based} approaches consider additional penalties or constraints during training to maintain the memory of old tasks. 
For example, Learning without Forgetting (LwF) exploits the idea of knowledge distillation \cite{hinton2015distilling} to force the model outputs on the current data to be similar to the old model's.
\\
Similarly, \textit{Pseudo-Label} also uses the model trained on the previous task \cite{guan2018learn}.
While visiting the data of a new task, the new task samples are passed through the old model, and the predictions on the old labels are added to the new ground truth targets.
\\
\textit{Architecture-based approaches}, as the name suggests, are methods that alter the original model's architecture to preserve previous knowledge. Techniques within this category employ various strategies for modifying the architecture \cite{rusu2016progressive, fernando2017pathnet, mallya2018packnet}. 
\\
Moreover, in CL literature, some methods propose combining different families of strategies to obtain better and more robust performance.
For example, Hou et al. \cite{Hou_2018_ECCV} combine a distillation-based approach with Replay. 
Another famous example is Dark Experience Replay (DER) \cite{buzzega2020darkexperiencegeneralcontinual}; rather than the ground truth targets, this method stores the logits produced from the previous models. In this way, the knowledge is distilled from the previously optimized models while replaying old input samples.

\subsection{CL in the Medical Domain}
\label{subsec:intro_medical_cl}
Previous works for CL in the medical domain have explored DIL and CIL scenarios within CL methods for classifying chest X-ray images. 
For the DIL scenario, Lenga et al. \cite{lenga2020continual} conduct a study considering two tasks based on chest X-ray images; Srivastava et al. \cite{srivastava2021continual} address a similar setting through a replay-based approach.

Singh et al. \cite{singh2023class} consider a CIL scenario with three tasks encompassing 12 classes, gradually introducing new classes over time. Similarly, Akundi et al. \cite{akundi2022incremental} study a CIL scenario with five tasks, each with only one class, for a total of 5 classes considered. 

In this setting, Akundi et al. \cite{akundi2022incremental} propose using distillation-based approaches.
However, previous works found a significant degradation in the performance of these approaches when assuming that certain classes encountered at the start of the data stream rarely reappear in subsequent tasks \cite{9575493}. 
These considerations bring additional challenges that previously proposed methods could not handle correctly. 

\section{Considered Scenario}
\label{sec:our_scenario}

\subsection{Motivation}
We consider the problem of flexible collaboration across medical institutions, enabling effective data pooling and model sharing. The nature of healthcare systems and processes leads to three challenges. First, medical data is highly sensitive and subject to stringent data protection regulations \cite{gostin2001national}. Sharing and processing medical data across multiple institutions is especially difficult as hospitals are independent organizations with distinct data controllers \cite{dove2018eu}. Informed consent can enable broader processing of medical data, but it is challenging to obtain for a large and representative group of patients \cite{dankar2019informed}.

Furthermore, medical diagnosis encompasses numerous disease labels, necessitating the curation and annotation of a comprehensive dataset across multiple pathologies, which can take several years \cite{kumar2018example}.
It is crucial for models to evolve incrementally and incorporate information about new diseases as they become available \cite{sun2023few} or surface, as was the case with COVID-19. 

Moreover, the classification of rare pathologies often benefits from extending models to related yet different conditions \cite{quellec2020automatic}. 
Typically, access to the original data used for training the model is not feasible due to data protection regulations. Consequently, a scenario where the model is updated with only a small subset of historical data, like the CIL setting, becomes crucial.

Finally, domain shift is inevitable, e.g., due to differences in medical equipment, acquisition protocols, and evolving populations \cite{lenga2020continual,karani2018lifelong,sirshar2021incremental}. This is especially true for datasets spanning multiple sites, which may serve areas with different demographics or provide specialized care, e.g., to different age ranges. In general, medical processes adopted at one hospital influence data acquisition and cause artifacts in the data \cite{ozgun2020importance}.

Overall, both DIL and CIL scenarios are common in the medical domain. Clearly, they do not represent mutually exclusive settings. Hence, a situation in which both new domains and new classes appear is not only possible but also very plausible, and it needs to be addressed.
For this reason, we consider a scenario termed \textbf{New Instances \& New Classes}. NIC combines the challenges of both new class arrivals and domain shifts within a single framework, mirroring the complexities often encountered in realistic applications like medical imaging. This combination of different types of shifts between tasks makes this scenario more challenging and suitable for representing intricate settings, such as the medical domain. 
The concept of the NIC scenario was initially introduced and investigated by Lomonaco et al. \cite{lomonaco2017core50}. 
It was examined within the context of a single-label classification problem, where it is assumed there are no intersections among tasks regarding labels or samples. To the best of our knowledge, this scenario was never analyzed in multi-label settings or the medical imaging setting.

\subsection{Scenario Description}
In this work, we model a NIC scenario in chest X-ray imaging across two sites. We consider the ChestX-ray14 dataset (NIH), compiled from patients of the NIH Clinical Center \cite{wang2017chestx} and the CheXpert dataset (CXP), with radiographies from patients treated at the Stanford Hospital \cite{irvin2019chexpert}. Both datasets contain information on 14 diseases, 7 of which are in common. We removed non-pathology labels from CXP, resulting in a stream of 19 classes. 
We divided CXP into three tasks and the NIH dataset into four tasks, ordering them sequentially to create the NIC stream of tasks represented in Figure \ref{fig:nic_scenario}. Each task models the availability of new data at one hospital. Between tasks $i$ and $i+1$, either new classes are introduced, or data from previously known classes is presented with a shift in the input data distribution. This shift occurs because new data on known classes becomes available from a different hospital with respect to previous tasks.

In each task, all and only the images containing the associated pathologies in the corresponding dataset are present.

Since several images contain multiple labels, the intersection among tasks is not null, presenting a realistic scenario where images with previously seen (but currently not labeled) diseases can appear in new tasks.

\subsection{Challenges}
\label{subsec:challenges}
The proposed benchmark considers a multi-label image classification problem where new classes and domains are added over time.
We identify these three challenges in the current literature landscape of multi-label classification:
(i) the task interference problem in replay approaches, 
(ii) the potential not exploited of the replay memory samples,
(iii) the strong forgetting suffered by distillation methods if previously seen classes don't reappear in the current task.

Indeed, while replay-based approaches have shown remarkable performance in multi-class classification problems \cite{yang2022benchmark}, they encounter a specific challenge known as \textit{task interference} in multi-label continual learning (CIL) settings. This interference arises between the current task samples and the replayed samples, when the intersection between tasks is not null, as each sample only contains information about the labels of the relative task.
For instance, if Task 1 is associated with the pathologies "Consolidation", "Edema", and "Effusion", and Task 2 with "Atelectasis" and "Hernia" a patient exhibiting both Consolidation and Hernia would be labeled "Consolidation" in Task 1 and "Hernia" in Task 2. Consequently, if this sample from Task 1 is selected for replay in Task 2, it would appear twice with two different targets.
The same problem is encountered when solving other CL problems in multi-label settings, such as Object Detection \cite{shmelkov2017incremental} and Semantic Segmentation \cite{9575493}. 

Moreover, we state that another problem of the Replay approach is that the replay memory is not fully exploited. Indeed, for each stored image, only the labels seen during its task are available, while all other labels are assumed to be unknown or absent.
However, this hides the potential of the memory, which could contain much more information than what is actually stored.

Distillation-based methods have demonstrated suboptimal performance in single-label classification tasks \cite{lenga2020continual}. Conversely, they find extensive application in other problems such as semantic segmentation \cite{akundi2022incremental} and object detection \cite{shmelkov2017incremental}.
However, distillation methods demonstrate significant forgetting when previously encountered classes do not reappear in the current task, a scenario often unrealistic in practical applications. In other words, unlike replay methods, distillation-based methods used in multi-label settings perform better when there is a considerate intersection between tasks.

In the considered scenario, all these challenges are present. There is some intersection between tasks, causing the interference issue in Replay. However, the diseases of the first tasks are rare, hence they do not appear frequently in subsequent tasks, thereby challenging distillation methods. Lastly, since the considered setting is multi-label, the problem of the under-exploitation of the replay buffer is present.


\begin{figure*}[!t]
  \centering
  \begin{subfigure}[b]{0.9\linewidth}
    \centering
    \fbox{
    \includegraphics[width=\linewidth]{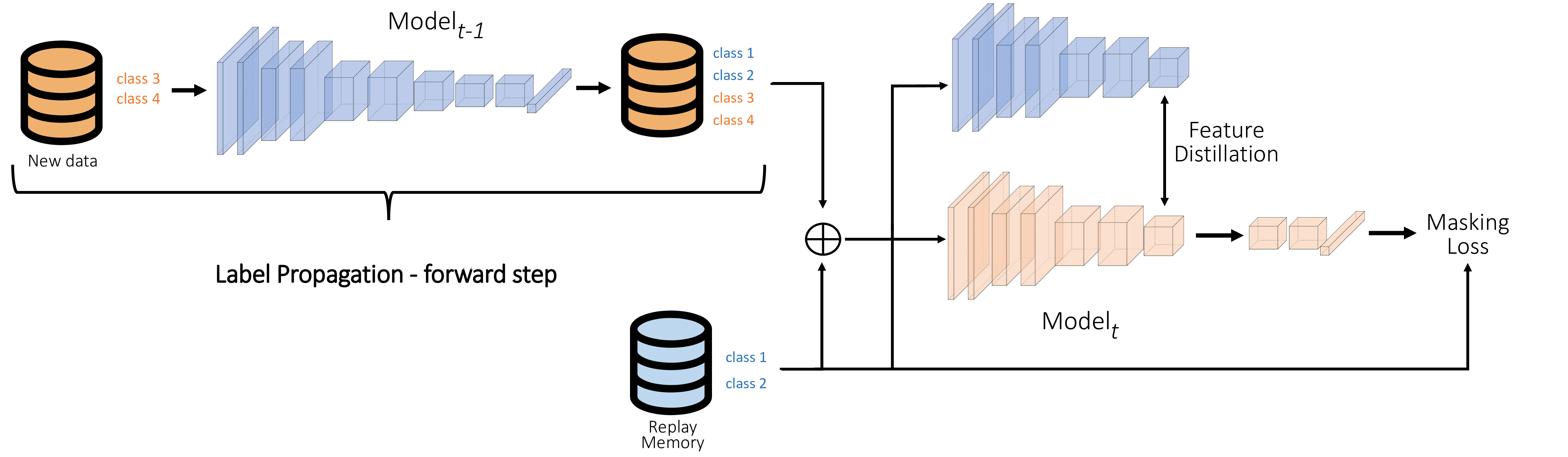}
    }
    \caption{Training Framework}
    \label{fig:training_framework}
  \end{subfigure}
  \hfill
  \begin{subfigure}[b]{0.9\linewidth}
    \centering
    \fbox{
    \includegraphics[width=\linewidth]{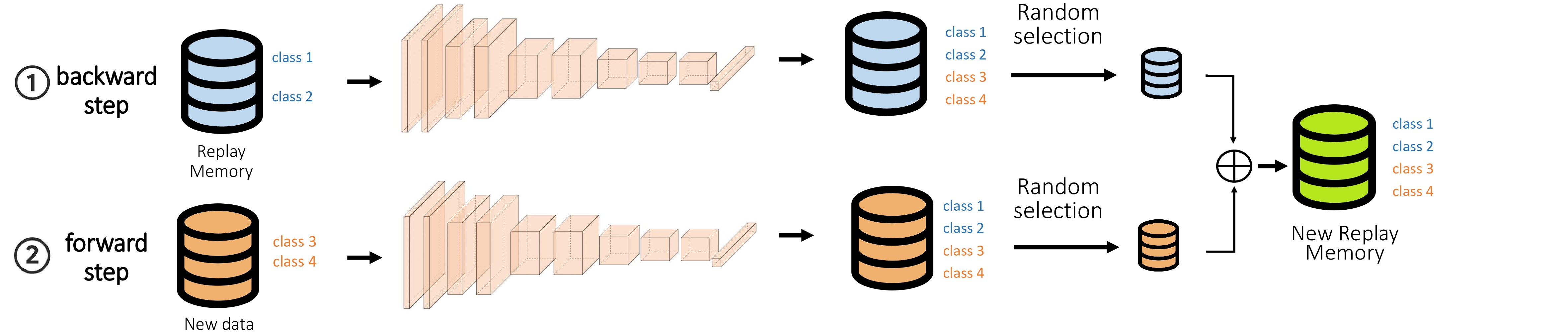}
    }
    \caption{Replay memory Consolidation with Label Propagation}
    \label{fig:RCLP}
  \end{subfigure}
  \caption{}
  \label{fig:OurApproach}
\end{figure*}
\section{Proposed approach: Replay Consolidation with Label Propagation}
\label{sec:our_approach}

To tackle the challenges described in Sec. \ref{subsec:challenges}, we propose Replay Consolidation with Label Propagation.
The advantages of our approach are threefold: (1) the integration of old knowledge on the new task samples and of new knowledge on the replay buffer samples, together with a Masking Loss, mitigate the issue of task interference of Replay; (2) the Replay memory is optimized since the targets provide information not only on the labels from their originating tasks but also from all preceding tasks up to the current one; (3) by replaying samples that contain old classes, the limitations of distillation are overcome.
The comprehensive strategy is depicted in Fig. \ref{fig:OurApproach}.

\subsection{Label Propagation: Forward step}

The first issue we address is that during the training of the model on new samples, these samples contain information only about the new labels and lack information on labels seen in previous tasks.
Therefore, in this part, we aim to incorporate the information on the old labels into the new data, and subsequently in the replay buffer, via the pseudo-labeling technique.

This step can be defined formally as follows.
Let $X_t, Y_t$ be the data associated with task $t$.
Let $X_M, Y_M$ be the data associated with the Replay memory.
Let $f_{\theta_t}$ be the model after being trained on task $t$.
Let $L_i$ be the set of labels associated with task $i$, $y^{L_i}$ the output produced by the model concerning the labels $L_i$, and $y_j^{L_i}$ the j-th output from the labels $L_i$.
During the training of task $t$, given a sample $x \in X_t$, the forward step adjusts the ground truth associated with each $y \in Y_t$, relative to the labels $L_1, \ldots, L_{t-1}$, to integrate the knowledge of the previously optimized model $f_{\theta_{t-1}}$.
To achieve this goal, we utilize the predictions generated by the previous model, denoted as $\hat{y}^{L_1}, \ldots, \hat{y}^{L_{t-1}}$. These predictions undergo a thresholding process where each class label is determined. Subsequently, these predicted labels replace the outdated labels in the ground truth vector $y$.

\begin{equation}
\forall y \in Y_t   ~~ y_j^{L_i} = \begin{cases}
1, & \text{if } \hat{y_j}^{L_i} > h \\
0, & \text{if } \hat{y_j}^{L_i} \leq h
\end{cases}
\quad
\begin{array}{c}
1 \leq j \leq |L_i| \\
1 \leq i \leq t-1
\end{array}
\label{eq:pseudo_labeling}
\end{equation}

After training on the new task $t$, a subset of the samples from task $t$ is saved in the memory buffer $M$. The targets associated with these samples, relative to old tasks labels, are determined using the label propagation technique previously described. This approach ensures that, following the training on task $t$, the memory buffer contains samples with information relevant to all tasks up to and including task $t$.

\begin{figure}[!h]
  \centering
  \begin{subfigure}[b]{0.33\linewidth}
    \centering
    \includegraphics[width=\linewidth]{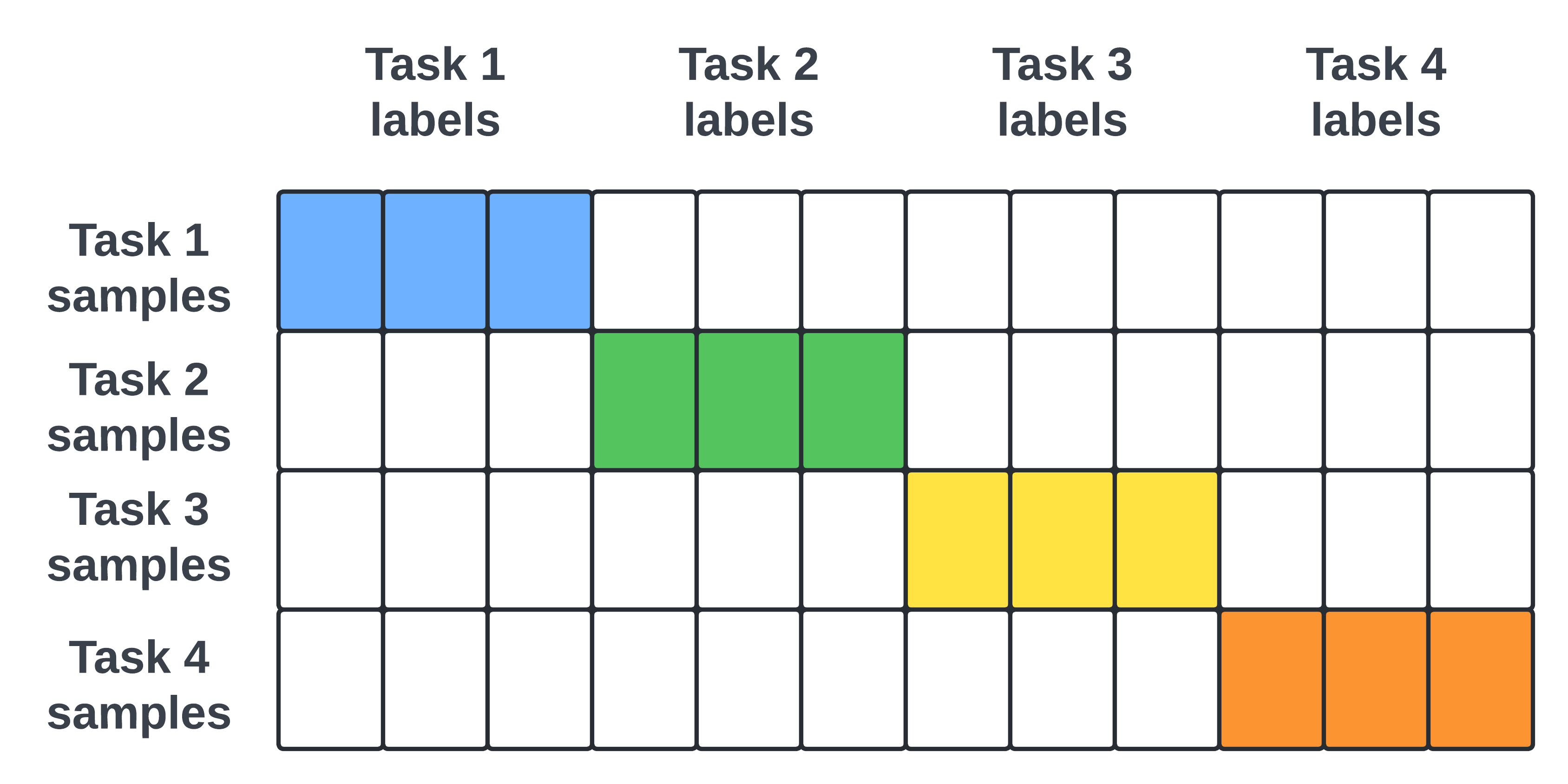}
    \caption{Replay Memory}
    \label{fig:replay_memory}
  \end{subfigure}
    \hfill
  \begin{subfigure}[b]{0.33\linewidth}
    \centering
    \includegraphics[width=\linewidth]{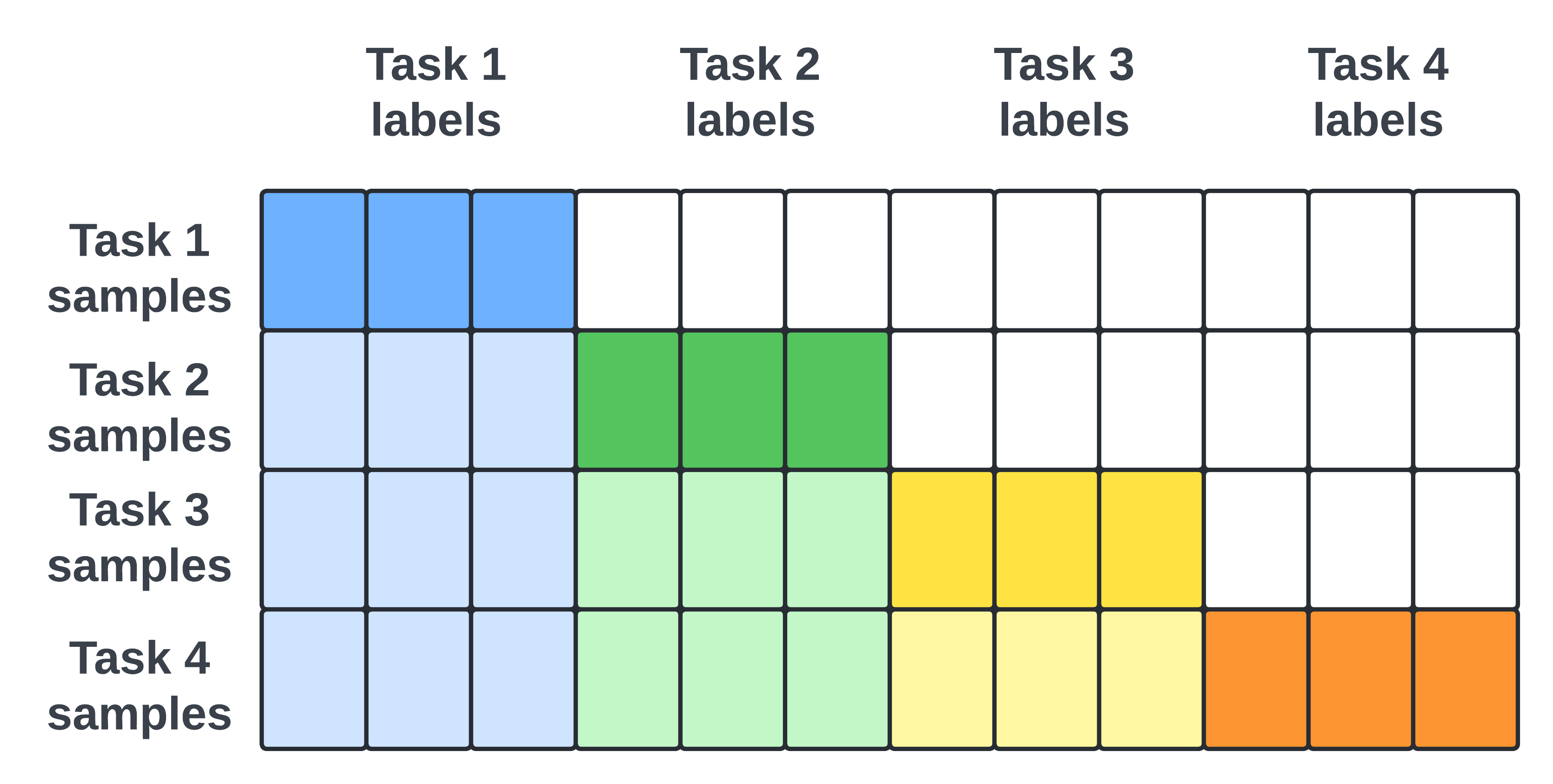}
    \caption{RCLP Forward step memory}
    \label{fig:RCLP_forward_memory}
  \end{subfigure}
  \hfill
  \begin{subfigure}[b]{0.33\linewidth}
    \centering
    \includegraphics[width=\linewidth]{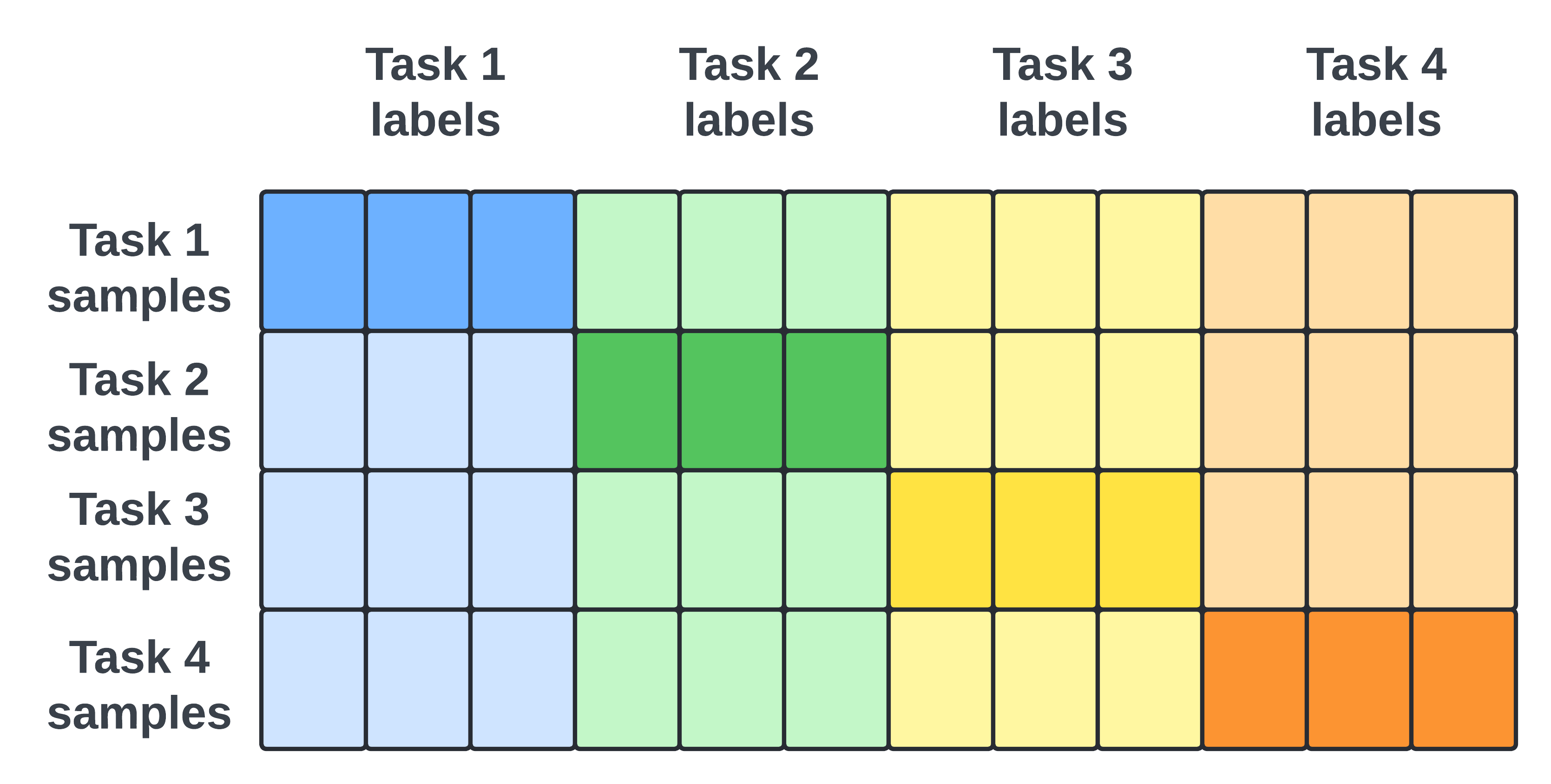}
    \caption{RCLP Memory }
    \label{fig:RCLP_memory}
  \end{subfigure}
  \caption{Representation of replay memories for different approaches. The vertical axis represents the samples of each task, while the horizontal axis is the labels associated with each task. (a) Each task of the Replay memory has information only on the labels seen during its iteration. (b) The forward step of label propagation, instead, saves samples that are informative of all tasks up to the current one. (c) Lastly, the backward step integrates the knowledge of the new labels in the replay buffer.
  }
  \label{fig:ComparisonMemory}
\end{figure}
\subsection{Label Propagation: backward step}
As described in section \ref{subsec:challenges}, the main problem of the replay approach is that the samples for each task contain only the labels of said task, as shown in Fig. \ref{fig:replay_memory}.
This represents an underuse of the potential information of the data contained in memory, which could be able to express much more, and it's a partial cause of the interference issue.
We partially solve this problem by performing the forward step of Label Propagation when creating the new replay memory.
In this way, samples from task $i$ in the memory buffer are informative on all tasks up to task $i$, as shown in figure \ref{fig:RCLP_forward_memory}.

While this approach reduces the interference issue, the samples in the memory buffer do not contain any information about the tasks that come after their respective task of origin.
Therefore, we propose the backward step of Label Propagation to consolidate the new knowledge acquired after training on a new task in the old samples of the replay buffer, as shown in figure \ref{fig:RCLP_memory}.

The procedure can be described formally as:
\begin{equation}
\forall y \in Y_M   ~~ y_j^{L_t} = \begin{cases}
1, & \text{if } \hat{y_j}^{L_t} > h \\
0, & \text{if } \hat{y_j}^{L_t} \leq h
\end{cases}
\quad
\begin{array}{c}
1 \leq j \leq |L_t|
\end{array}
\end{equation}

In other words, after training the model on the new task $t$, the samples in the replay buffer, coming from previous tasks, are given as input to the model, and the pseudo-labels relative to the current task are computed and added to the ground truth targets. 

\subsection{Masking Loss}
After performing the label propagation step, only the labels relative to the current task in the replayed samples are absent. 
Therefore, we propose a masking loss to overcome this issue.
When the model revisits old samples, the masking loss ensures that the model's outputs are only influenced by the labels of the old tasks. This effectively 'masks' the influence of the old samples on the new labels, preventing task interference.
\begin{equation}
    L_{M}(x,y,\hat{y})= \begin{cases} L_{BCE}(y,\hat{y}), & x \in X_t \\ \sum_i^{t-1} L_{BCE}(y^{L_i},\hat{y}^{L_i}), & x \in X_M\end{cases}
\end{equation}

\subsection{Feature Distillation}
To further improve the model's performance, we consider the feature distillation technique on the intermediate representations of the model \cite{li2024continual}.
Let our model $f$ be composed as $g_\phi(h_\omega(x))$, where given $x$, $h$ produces the features, and the classifier $g$ is trained on such features.
The main idea of feature distillation is that the features produced by the new model on the new and old data should not deviate too far from the features produced by the previous model on said data.
This allows the model to maintain a more consolidated memory mitigating the effects of catastrophic forgetting.
Formally, given an input sample $x$:
\begin{equation}
    L_{FD} = || g_{\phi_t}(h_{\omega_t}(x)) - g_{\phi_{t-1}}(h_{\omega_{t-1}}(x)) ||
\end{equation}
The input could either come from the new task or from the replay buffer.
Therefore the final loss is built in the following way:
\label{eq:RCLP}
\begin{equation}
    L(x,y,\hat{y}) = L_M(x,y,\hat{y}) + \gamma \cdot L_{FD}(x)
\end{equation}

\subsection{Method Analysis}
In conclusion, RCLP optimally leverages the strengths of both distillation and rehearsal methods.
To address the problem of task interference, we split our RCLP approach into multiple components. 
We propose the Label Propagation technique, which can be divided into two steps.
The \textbf{forward step}, like Pseudo-Label, enables new samples to leverage the knowledge from a previously optimized model to acquire the old labels. However, unlike Pseudo-Label, our approach applies this step not only during training but also to consolidate information within the replay memory buffer.
Instead, the \textbf{backward step} allows the injection of the new knowledge (acquired after the training on the new task) into the samples of replay memory.
In this way, a relevant advantage of Label Propagation is represented by the full exploitation of Replay memory by utilizing samples that contain information on all seen tasks (as depicted by Fig. \ref{fig:ComparisonMemory}).
Another advantage is that the injected labels are inserted into the replay memory and remain unchanged. This provides more stable information compared to the dynamically generated labels in methods like LwF, LwF-Replay, and Pseudo-Label.
Moreover, we introduce a \textit{Masking Loss} to further mitigate the task inference issue by considering only the old labels in the replay samples, hence avoiding the calculation of loss on the labels of the current task.
It should also be noted that our approach does not require additional memory compared to Replay, as the method only fills empty labels with pseudo-labels within the existing replay buffer.
In addition, to enhance final performance and reduce further forgetting, a feature distillation loss is implemented.
\section{Experimental Setting}
 
\label{sec:experimental_setting}
We test the following methods against the benchmark: Joint training, Fine tuning, Replay, LwF, Pseudo-Label, LwF-Replay, DER.
Following the literature on training classification models on chest X-ray images \cite{seyyed}, we selected a 121-layer DenseNet \cite{huang2018denselyconnectedconvolutionalnetworks} architecture pre-trained on ImageNet \cite{5206848} for our study. We employed the binary cross-entropy loss function and optimized the models using the Adam \cite{kingma2017adammethodstochasticoptimization} optimizer with a learning rate set at 0.0005. For both datasets, we used an 80-10-10 split between training, validation, and test sets. Consistent with findings from previous works \cite{weng2023sexbased} indicating that using a single image per patient does not significantly impact performance, we included only one image per patient, prioritizing frontal images. All images were resized to 256 $\times$ 256 and normalized via the mean and standard deviation of the ImageNet dataset.
As an upper bound for our methods, we considered Joint Training, which consists in simultaneously training on all tasks. In this case, we reduced the learning rate by a factor of 2 if the validation loss did not improve over three epochs and stopped training if the validation loss did not improve over 10 epochs. 
For all other methods besides Joint training, we trained on each task for 10 epochs.
As a lower bound, we evaluated the Fine-Tuning approach, where each task is trained sequentially, with no additional technique to avoid catastrophic forgetting. For the Replay approach, we consider a mix ratio of 50\% and a memory size corresponding to 3\% of the original dataset size.
In LwF \cite{li2017learning}, the loss is computed as $L = L_1 + \tau L_2$, where $L_1$ is the loss for the current task, while $L_2$ is the distillation loss; we set $\tau$ = 2, following the literature \cite{li2017learning}.
For Pseudo-Label, the threshold hyperparameter is extremely important. We define a different threshold for each class, choosing the one found to maximize the F1 score on the validation sets relative to the origin task of each label. When no new labels are introduced between task $i$ and task $i+1$, the thresholds remain fixed at the values that were optimal in the previous validation sets.
For the LwF-Replay approach, we used the same hyperparameters of Replay and LwF.
Similarly, for the DER approach, the same hyperparameters of Replay are used.
Finally, for our approach, we adopt the same hyperparameters as in Replay and employ the same strategy for computing thresholds as in Pseudo-Label.
Moreover, for computing the intermediate distillation loss, we utilized the output of the 12th block out of the 16 blocks in the DenseNet architecture. Lastly, the parameter $\gamma$ of the feature distillation loss in Eq. \ref{eq:RCLP} is set to 1.

\textbf{Evaluation metrics}
We use the macro F1 score as the primary metric to evaluate model performance since it is a common metric when dealing with multi-label classification problems \cite{ge2018multi,wang2016cnn}.
The macro F1 is the average F1 score calculated for each label. The F1 score quantifies the balance between precision and recall, hence it is an appropriate choice for the evaluation of imbalanced datasets. Its focus on positives, i.e., images where a given pathology is present, makes F1 especially relevant in the medical domain. We additionally report the AUC ROC, a frequently used metric for evaluating models trained to classify medical images \cite{seyyed, lenga2020continual, srivastava2021continual}. The AUC is a valuable metric for evaluating the overall discriminative power of a classification model.
However, it is not considered sufficient for datasets with skewed class balance, such as the considered datasets. Indeed, the resulting value may be misleading, hiding the poor performance of a model \cite{jeni2013facing}. 

For the evaluation over a stream of tasks, we consider the Average F1 over all the pathologies of all tasks up to $i$, as typical in CL papers \cite{akundi2022incremental, lenga2020continual}. 
Similarly, we measure the percentual forgetting, proposed in previous papers in CL literature \cite{dalle2023multi, kim2020imbalanced}, hence the percentage difference between the final performance of the model on each task and the performance just after training on said task. Lastly, we consider the Relative gap between each method's final performance and the upper baseline's performance, i.e., Joint Training.

\begin{figure}[!th]
   \centering
   \includegraphics[width=0.8\linewidth, trim = 0 0 0 0]{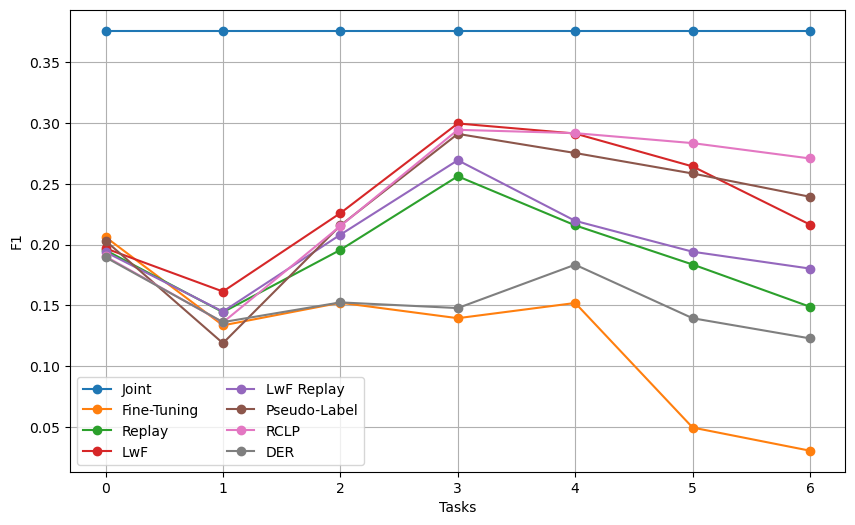}
   \caption{F1 score during the task stream for each method.}
   \label{fig:f1_performance}
\end{figure}

\begin{table*}[t]
\caption{In \textbf{bold} is reported the best value for each metric among the compared CL approaches. 
Average F1, Forgetting, and Relative gap metrics are as defined in Sec. \ref{sec:experimental_setting}.}
\label{tab:results}
\centering
\begin{tabular}{|c|ccc|c|}
\hline
\textbf{}           & \multicolumn{4}{c|}{\textbf{Metrics}}                                                       \\ \hline
\textbf{Strategy}   & \multicolumn{1}{c|}{\textbf{Avg. F1} $\uparrow$} & \multicolumn{1}{c|} {\textbf{Avg. AUC} $\uparrow$} & \multicolumn{1}{c|}{\textbf{Forgetting F1 } $\downarrow$} & \textbf{Relative gap $\downarrow$ } \\ \hline
\textbf{Joint Training}      & \multicolumn{1}{c|}{0.38}            & \multicolumn{1}{c|}{0.79}               &      -        & -        \\ \hline
\textbf{Fine-Tuning }        & \multicolumn{1}{c|}{0.02}            & \multicolumn{1}{c|}{0.55}               &      100\%      & 95\%    \\ \hline \hline
\textbf{Replay }             & \multicolumn{1}{c|}{0.15}            & \multicolumn{1}{c|}{0.65}               &      59\%    & 60\%  \\ \hline
\textbf{LwF}                 & \multicolumn{1}{c|}{0.22}            & \multicolumn{1}{c|}{0.68}               &      41\%    & 42\%  \\ \hline
\textbf{LwF Replay }         & \multicolumn{1}{c|}{0.18}            & \multicolumn{1}{c|}{0.68}               &      48\%    & 52\%  \\ \hline
\textbf{DER}       & \multicolumn{1}{c|}{0.12}            & \multicolumn{1}{c|}{0.58}               &      73\%    & 67\%  \\ \hline
\textbf{Pseudo-Label}        & \multicolumn{1}{c|}{0.24}            & \multicolumn{1}{c|}{\textbf{0.69}}            &      21\%    & 36\%  \\ \hline
\textbf{RCLP (ours) }& \multicolumn{1}{c|}{\textbf{0.27}}   & \multicolumn{1}{c|}{\textbf{0.69}}               &      \textbf{2.4\%}     & \textbf{27\%} \\ \hline
\end{tabular}
\end{table*}

\section{Results}
\label{sec:results}
In the next part, we discuss the outcomes produced by each method when applied to the proposed challenging benchmark. Specifically, we will elucidate the achieved results concerning the F1 performance in Sec. \ref{subsec:performance_results}. Subsequently, we delve deeper by examining the forgetting behavior exhibited by each method in Sec. \ref{subsec:forgetting_results}.

\subsection{Performance results}
\label{subsec:performance_results}

The results are reported in Tab. \ref{tab:results}, while the average F1 metric over time for each method is shown in Fig. \ref{fig:f1_performance}. The reported results are averaged over multiple repetitions to reduce the effect of randomness on
the experimental results.

Despite the issue of task interference, Replay has an average F1 score of 0.15, still much higher than the lower bound Fine-Tuning that achieves an F1 of 0.02. 
This discrepancy in performance between the two methods can be observed from the relative gap as well: 95\% for Fine-Tuning and 60\% for Replay.
LwF outperforms Replay, as expected from the literature \cite{shmelkov2017incremental}, exhibiting a significant improvement, achieving an average F1 score of 0.22.
Pseudo-Label performs similarly to LwF, with a marginally higher F1 score of 0.24.

Interestingly, the combination of LwF and Replay, LwF-Replay, isn't able to produce satisfactory results, obtaining an F1 value (0.18) lower than that of the LwF and Pseudo-Label methods taken singularly.
This is expected due to the interference caused by Replay, which is not overcome by combining the method with LwF. For the same reasons, DER performs very poorly as well, achieving a final F1 of 0.12, lower than all methods besides Fine-Tuning. 
Indeed, as discussed, a series of challenges need to be addressed to exploit the advantages of replay-based methods together with distillation-based methods in the multi-label setting.
Lastly, our approach, Replay Consolidation with Label Propagation, achieves an F1 score of 0.27 and a relative gap of 27\%, exhibiting a notable improvement with respect to the other methods, as shown in Tab \ref{sec:results} and Fig. \ref{fig:f1_performance}).

\begin{figure}[!th]
   \centering
   \includegraphics[width=0.8\linewidth, trim = 0 0 0 0]{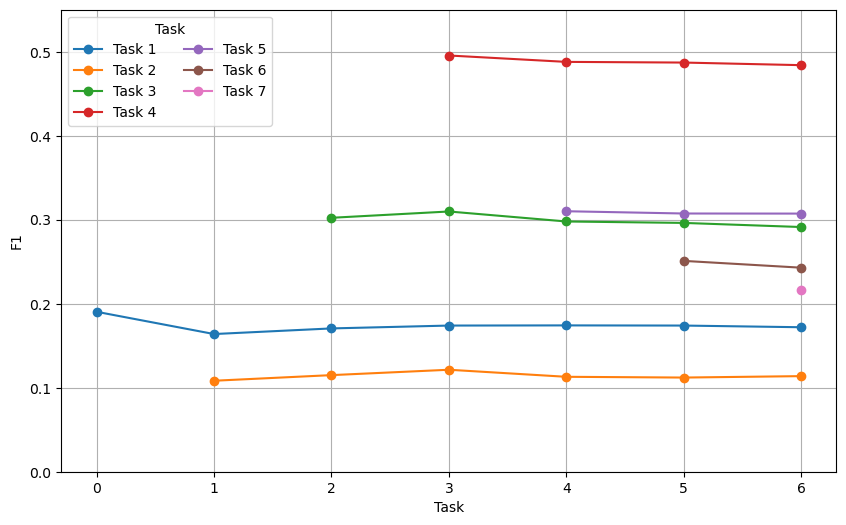}
   \caption{Performance for each task over time of the RCLP approach. The behavior of each curve is very stable and the related forgetting value is very low (2.4\%).  }
   \label{fig:forgetting_RCLP}
\end{figure}

\subsection{Forgetting Analysis}
\label{subsec:forgetting_results}
Another interesting analysis concerns the forgetting metric (shown in Tab. \ref{tab:results}).
The first general insight is that the forgetting values are notably high across all methods besides RCLP, ranging from 73\% for DER to 21\% for Pseudo-Label.
When examining our proposed approach, RCLP, we note a remarkably low forgetting rate of 2.4\%, with the final performance clearly outperforming all other methods.
This is also supported by Fig. \ref{fig:forgetting_RCLP} showing the performance over time of each task when considering the RCLP approach. This plot displays RCLP's excellent performances with stable F1 curves for each task.

\section{Conclusions and Future Work}
\label{sec:conclusions_future_work}
Our work introduces a novel benchmark tailored for evaluating CL methods in the domain of multi-label medical image classification. 
This benchmark includes different medical imaging datasets, pathologies, and imaging conditions.
In particular, we designed the benchmark by considering realistic conditions in the medical settings and combining new classes and domains in the task stream.

Because of the challenges encountered in the scenario, we proposed a novel method called Replay Consolidation with Label Propagation (RCLP).
In particular, to address the problem of \textit{task interference}, a \textit{Masking Loss} is introduced.
Moreover, to better exploit the data contained in the replay memory, we propose a Label propagation technique composed of a backward and a forward step.
Our method outperforms existing approaches in the field of multi-label image classification within the medical domain, demonstrating significantly superior performance with minimal forgetting.

While our approach has demonstrated high effectiveness in handling multi-label classification problems within the medical setting, its applicability to other settings still needs to be explored. 
In future work, we plan to evaluate the effectiveness of RCLP in other scenarios, such as object detection and semantic segmentation problems.

\bibliographystyle{unsrt}  
\bibliography{references}  

\newpage

\section*{Appendix}

\subsection{Distribution}
In Fig. \ref{fig:nic_scenario}, we present the visual representation of the frequency of each pathology in each task. 
The blue bars correspond to the pathologies associated with the current task, while the light blue bars correspond to the other pathologies, and the blue contour represents the frequency of each disease in the original dataset. During each task, we keep all the images in the dataset containing at least one of the pathologies associated to the task; however, other diseases may be present even though the information on the presence of such pathologies is not available, hence they’re hidden pathologies. As can be noticed, there's some intersection between tasks; however, some diseases such as those of task 1 are very rare, hence they rarely appear in subsequent tasks.

\begin{figure*}[!h]
   \centering
   \includegraphics[width=\linewidth, trim = 0 0 0 0]{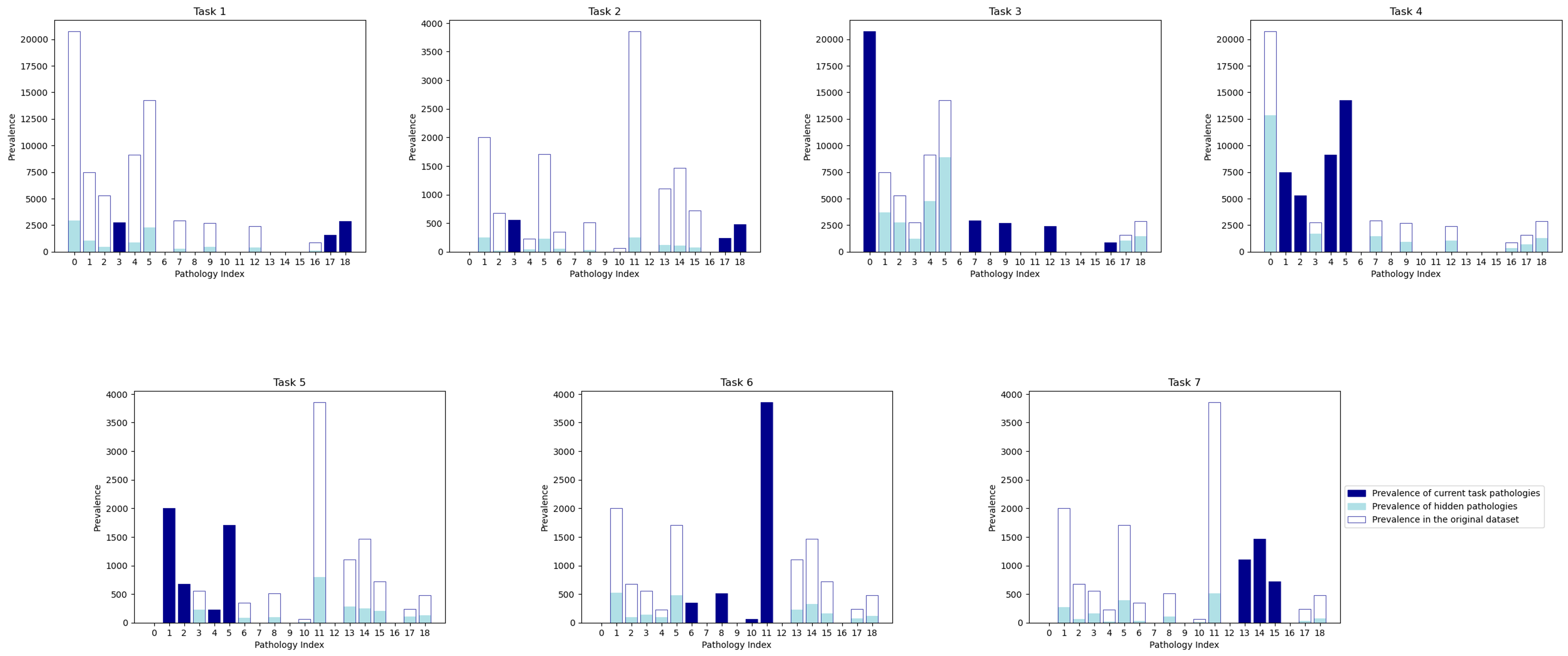}
   \caption{Visual representation of the frequency of each pathology in each task}
   \label{fig:nic_scenario}
\end{figure*}

\subsection{Results}

\begin{figure*}[t]
  \centering
  \begin{subfigure}[b]{0.49\linewidth}
    \centering
    \includegraphics[width=\linewidth]{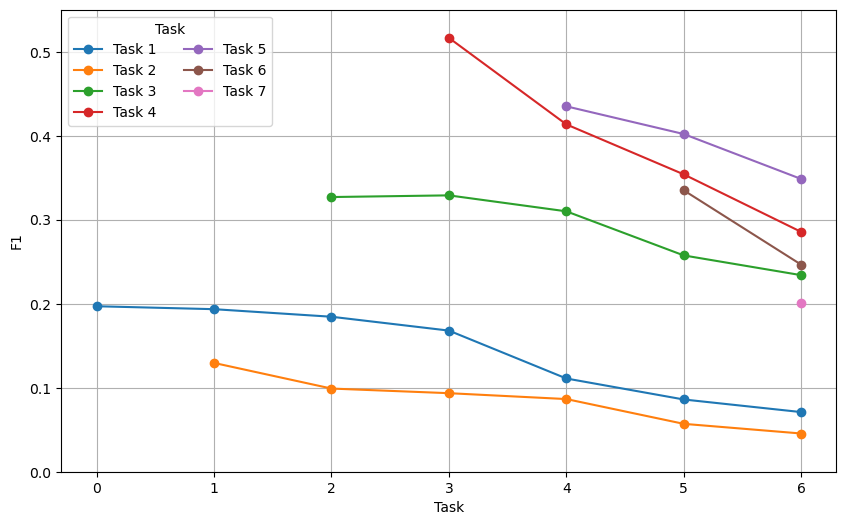}
    \caption{Learning without Forgetting (LwF)}
    \label{fig:forgetting_f1_LwF}
  \end{subfigure}
  \hfill
  \begin{subfigure}[b]{0.49\linewidth}
    \centering
    \includegraphics[width=\linewidth]{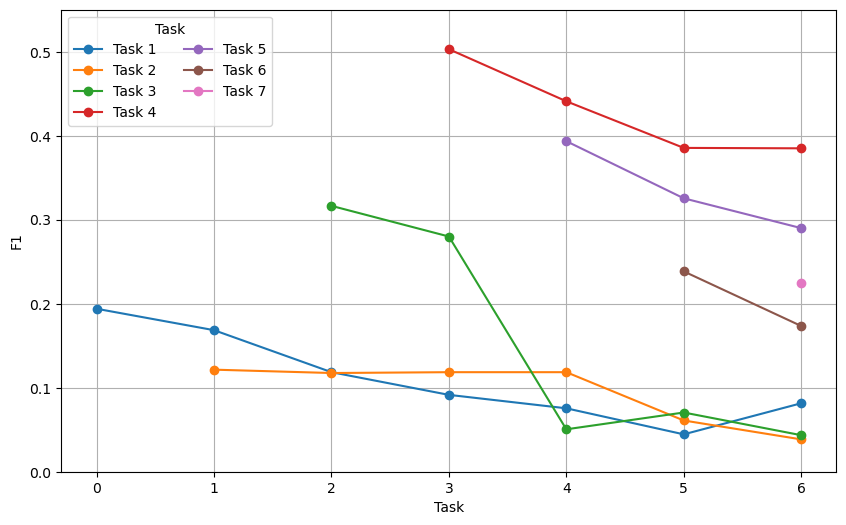}
    \caption{Learning without Forgetting (LwF) Replay}
    \label{fig:forgetting_f1_LwF_Replay}
  \end{subfigure}
  
  \medskip
  
  \begin{subfigure}[b]{0.49\linewidth}
    \centering
    \includegraphics[width=\linewidth]{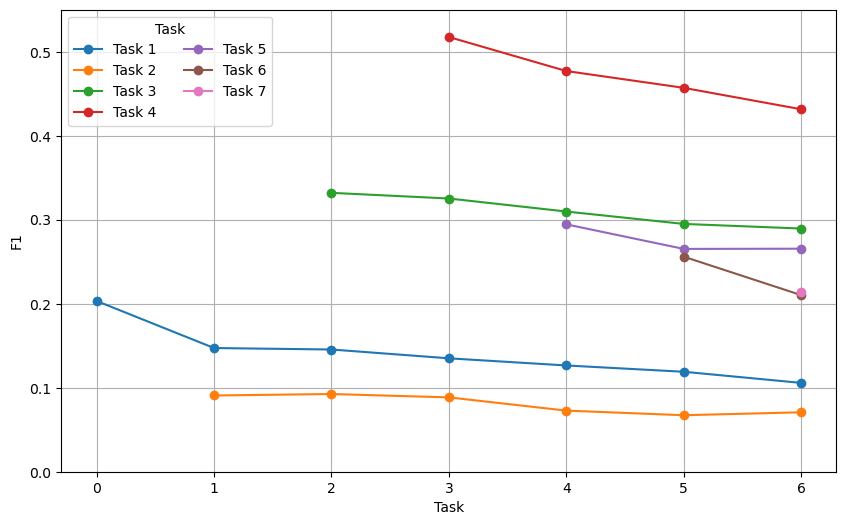}
    \caption{Pseudo-Label}
    \label{fig:forgetting_f1_pseudolabel}
  \end{subfigure}
  \hfill
  \begin{subfigure}[b]{0.49\linewidth}
    \centering
    \includegraphics[width=\linewidth]{f1_RCLP.png}
    \caption{RCLP}
    \label{fig:forgetting_f1_PLR}
  \end{subfigure}
  
  \caption{Comparison between the average F1 score on each task between RCLP and prior methods.}
  \label{fig:forgetting_plots}
\end{figure*}

In Fig. \ref{fig:forgetting_plots} the average F1 across all tasks after training on each task is presented, for our proposed method and three comparative methods: LwF, LwF Replay, and Pseudo-Label. Fig. \ref{fig:forgetting_f1_LwF} and \ref{fig:forgetting_f1_LwF_Replay} illustrate that both LwF and LwF Replay exhibit significant performance degradation on previous tasks after training on new tasks. The Pseudo-Label method demonstrates a substantial reduction in forgetting, which is further minimized with our RCLP method, as can be seen from Fig \ref{fig:forgetting_f1_pseudolabel} and \ref{fig:forgetting_f1_PLR}. Indeed, the curves of Fig. \ref{fig:forgetting_f1_PLR} are remarkably stable, demonstrating an overall forgetting rate of just 2.4\%.

\end{document}